# Adaptive Robot Perception in Construction Environments using 4D BIM

## Mani Amani.,[1] and Reza Akhavian, Ph.D.[2]

[1]DiCE Lab, Computational Science Research Center, San Diego State University; e-mail: mamani5250@sdsu.edu
[2]DiCE Lab, Department of Civil and Environmental Engineering, San Diego State University; e-mail: rakhavian@sdsu.edu

## ABSTRACT

Human Activity Recognition (HAR) is a pivotal component of robot perception for physical Human Robot Interaction (pHRI) tasks. In construction robotics, it is vital that robots have an accurate and robust perception of worker activities. This enhanced perception is the foundation of trustworthy and safe Human-Robot Collaboration (HRC) in an industrial setting. Many developed HAR algorithms lack the robustness and adaptability to ensure seamless HRC. Recent works have employed multi-modal approaches to increase feature considerations. This paper further expands previous research to include 4D building information modeling (BIM) schedule data. We created a pipeline that transforms high-level BIM schedule activities into a set of low-level tasks in real-time. The framework then utilizes this subset as a tool to restrict the solution space that the HAR algorithm can predict activities from. By limiting this subspace through 4D BIM schedule data, the algorithm has a higher chance of predicting the true possible activities from a smaller pool of possibilities in a localized setting as compared to calculating all global possibilities at every point. Results indicate that the proposed approach achieves higher confidence predictions over the base model without leveraging the BIM data.

## INTRODUCTION

Construction worker activity recognition is one of the main components in enabling Human-Robot Collaboration (HRC). Many previous studies utilize different modalities to enable HAR. Some works approach activity recognition through wearable sensors such as IMU (Akhavian et al., 2016; Xefteris et al., 2022), visual inferences (Beddiar et al., 2020), and using multiple modalities (Qi et al., 2022).

However, due to the dynamic nature of construction job sites, these models often fail to use robust, out-of-training data inferences. These models could be trained solely on construction activity tasks and data to improve robustness. Nevertheless, this approach is currently not feasible due to limited visual construction activity data. Another approach is to increase the depth and complexity of the models. Previous works have shown that deeper models can show better performance (Jha et al., 2021). However, this can result in higher inference times, creating a processing latency detrimental to real-time HRC.

One method of increasing accuracy can be reducing the *hypothesis space* and *label space* of the model. A hypothesis space in the context of machine learning (ML) is the entire set of possible models or functions that a neural network can represent, given its architecture. However, reducing the hypothesis space of neural networks and complex, deep models can be a very cumbersome task (Szymanski et al., 2018). Alternatively, one can attempt to reduce the label space of the models. Previous research has attempted to reduce the label space with the improvement of reducing computational complexity (Moyano et al., 2022). It is important to note that hypothesis space and label space are different but related terms. The hypothesis space presents the entire possible representations given by the model. The label space is simply the possible output that the model can give. By restricting the label space, we are, in effect, restricting the hypothesis space. In other words, the label space is a subset of the hypothesis space which covers the broader possibilities in each model. While the related work has focused on hypothesis space reduction to improve computational complexity. We aim to improve the robustness and accuracy of the algorithms. Therefore, this paper proposes to use deterministic 4D building information modeling (BIM) data to limit the subspace. By using pre-defined activities depending on the work being done during that date, we can improve the model performance by removing them from the label space. We hypothesize that this label space reduction will both improve the overall accuracy of the model and increase the confidence of the model in its predictions. The main contribution of this approach is to force pretrained models to have more robust use cases without the need for extensive retraining and dataset collection. This is crucial since construction-specific datasets and architectures are scarce. Nevertheless, there are many rich, deterministic information sources such as BIM that can help researchers to improve robotic deployment.

**METHODOLOGY**

HAR models can be extremely inaccurate when they classify different activities that result in similar patterns in the collected data, especially if there has not been sufficient data on each activity in the training dataset. For example, the tasks of *using a drill* and *using a nail gun* have very similar high-level activity profiles which can be even difficult for humans to differentiate at times. The similarity of the tasks can cause ML models to have an increasingly hard time differentiating between the two. We introduce the construction schedule to assist the model in its predictions. For example, a task such as installing drywall anchors will not need the usage of nail guns. We can reduce the label space by removing all of the activity classes unrelated to the scheduled task. X-CLIP's architecture allows us to modify the label space, in this case, the text classes, with simplicity. This gives us and any engineers in an applied setting complete flexibility in controlling the label space. Figure 1 shows the overall structure of the framework.

We utilize a pre-trained X-CLIP model (Ni et al., 2022) as our base model HAR model. The authors have previously investigated this model architecture for increasing the robustness of activity recognition in construction contexts by the authors before (Shahnavaz et al., 2023). The video classes encompass the entirety of possible construction-related activities that are done inside the work environment given by the 4D BIM. Since 4D BIM models have high-level task descriptions, we must break down these tasks into smaller activity bursts that together define the high-level task. We utilize a BIM pipeline that extracts the task from the BIM schedule and selects a predefined activity subset that aids the HAR model in reducing the label space. The pipeline utilizes the construction schedule to identify activities and cross reference with previously defined subtasks. Once those subtasks are dictated the label space will be truncated. The authors recorded and labeled the data using the BIM data, project manager's input, and the ongoing work at the job site.

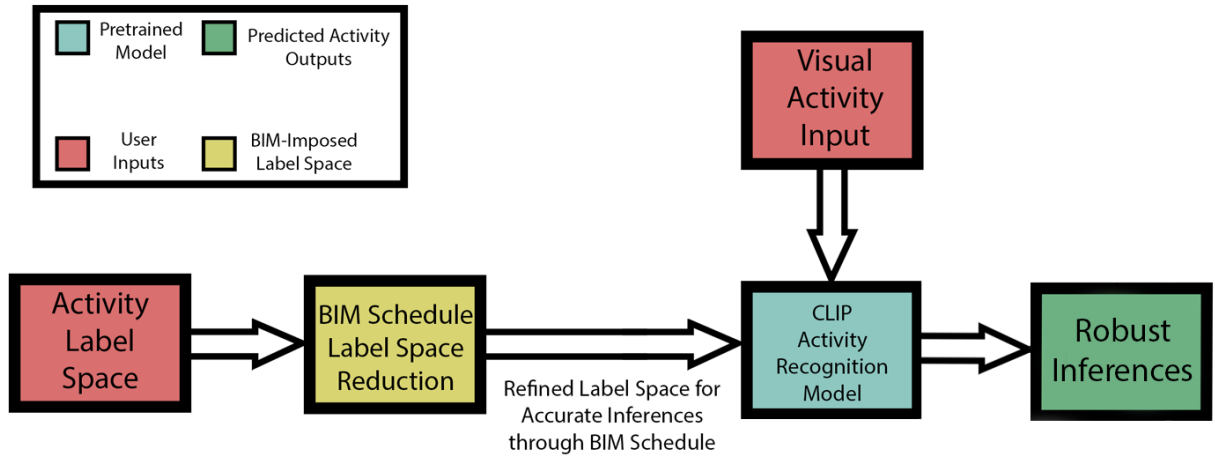

**Figure 1. Schematic of BIM Schedule Integration with X-CLIP for HAR**

The X-ClIP model calculates a similarity score between the text video embeddings through each respective encoder. These embeddings are achieved by passing through our text and video into each respective encoder. A transformer architecture then aggregates the predictions of each video frame. A softmax function can be employed to ascertain the probabilities of each task. The maximum likelihood can then be used as the highest confidence prediction. The number of classes the model is comparing with the video input directly impacts the number of potential activities that can be chosen.

Using the 4D BIM's schedule, we can extract what activities are possible given a specific predefined task. For example, if the task defined by the 4D BIM is "Painting a wall", we can reasonably infer that "Cleaning the wall with a brush" would not be one of the possible activity options. In this paper, we only analyze the effects of integrating the schedule component of 4D BIM with activity recognition. In future works, we plan to incorporate other components of 4D BIM such as localization and construction progress for robot development.

**EXPERIMENTAL RESULT**

Video data of construction activities was collected at a Housing construction project in San Diego. Two floors were designated by the project managers as having the most activity on that date and the authors collected the data. We filmed and partitioned the data depending on the schedule. Two Tasks were filmed:

- Assembling Metal Frames
- Installing Drywall

The Tasks had 5 and 6 distinct activities respectively. For the most part, these activities are not mutually exclusive. For example, both assembling metal frames and installing drywall involve measuring and drilling. The number of unique activities between both tasks was 7, giving the label space of our model 7 activities to choose from. To further simulate a realistic implementation of this framework, we added 11 more unrelated construction activities expected to happen on a job site. The label space for HAR without schedule integration consisted of 18 distinct activities being inferred by the ML model. However, with BIM integration we limit that label space to 5 and 6 tasks for the two Tasks, respectively. The tasks are described in the following:

**Table 1. List of Activities in Each Task**

| Label Spaces | Labels |
| --- | --- |
| **Entire Label Space** | Clamping, Grinding, Drilling, Measuring, Marking, Cutting, Nail gunning, Sawing, Picking up Trash, Shoveling, Using a Screwdriver, Hammering, Mixing Cement, Driving, Blowtorching, Laying Bricks, Soldering, Painting |
| **Task 1** | Clamping, Grinding, Drilling, Measuring, Marking, Cutting |
| **Task 2** | Drilling, Measuring, Marking, Cutting, Nail gunning |

We investigated the effect of the integration of the BIM schedule on the model in two different metrics: (1) precision performance metrics including Accuracy, Precision/Recall, and F1 score, and (2) confidence through the percentage of the maximum likelihood being true.
We evaluate the model for each task separately using the entire label space and one of the BIM-restricted label spaces for each task.

We chose to conduct these experiments with the baseline X-CLIP model available on Hugging Face (Hugging Face, 2023). CLIP and by extension, X-CLIP utilizes a training regimen using an infoNCE such as Equation 1 (Radford et al. 2021):

$$\mathcal{L} = -\frac{1}{N}\sum_{i=1}^{N} \log \frac{\exp(\text{sim}(x_i,y_i)/\tau)}{\sum_{j=1}^{N} \exp(\text{sim}(x_i,y_j)/\tau)} \qquad \text{Equation (1)}$$

X-CLIP training regimen contributes to our decision to use this as our model since previous works have shown that contrastive training can show increased robustness and generalizability (Parulekar et al., 2023). The base X-CLIP model has not been trained on construction-specific contexts. This will result in worse performance in out-of-training data inferences. While previous works have shown that fine-tuning the model with construction activity data can improve performance (Shahnavaz et al., 2023), we chose to proceed with the baseline model for a more objective and reproducible comparison. The architecture of the X-CLIP model incorporated with the BIM schedule funnel is depicted in Figure 2.

By applying the label space reduction by incorporating the schedule we were able to decrease the number of classes by 6-7 depending on the task. This truncation forces the model to process the probability space in its entirety without having additional classes. The result is a higher confidence scoring and fewer false predictions courtesy of removing factually wrong or fringe cases from the computations. Since the ML algorithm must calculate the probabilities from scratch, the distribution is still from 100%. The confidence percentages are higher than before since there are fewer classes to participate in the distribution, thus each current class will have a higher probability distribution. From there on, we evaluate the efficacy of the model using the combined and reduced label spaces. As shown in Table 2 incorporating BIM metrics positively influences the majority of precision scores. In addition to accuracy, we analyze precision, recall, and F1 scores for further validation of our framework's effects. These scores further shed light on the model's performance regarding false positives and negatives.

*Confidence* is referred to as the highest probability value among those assigned to the predicted classes by the model. This can be an important metric to enhance transparency and trust in the functional deployment of collaborative robots and intelligent models.

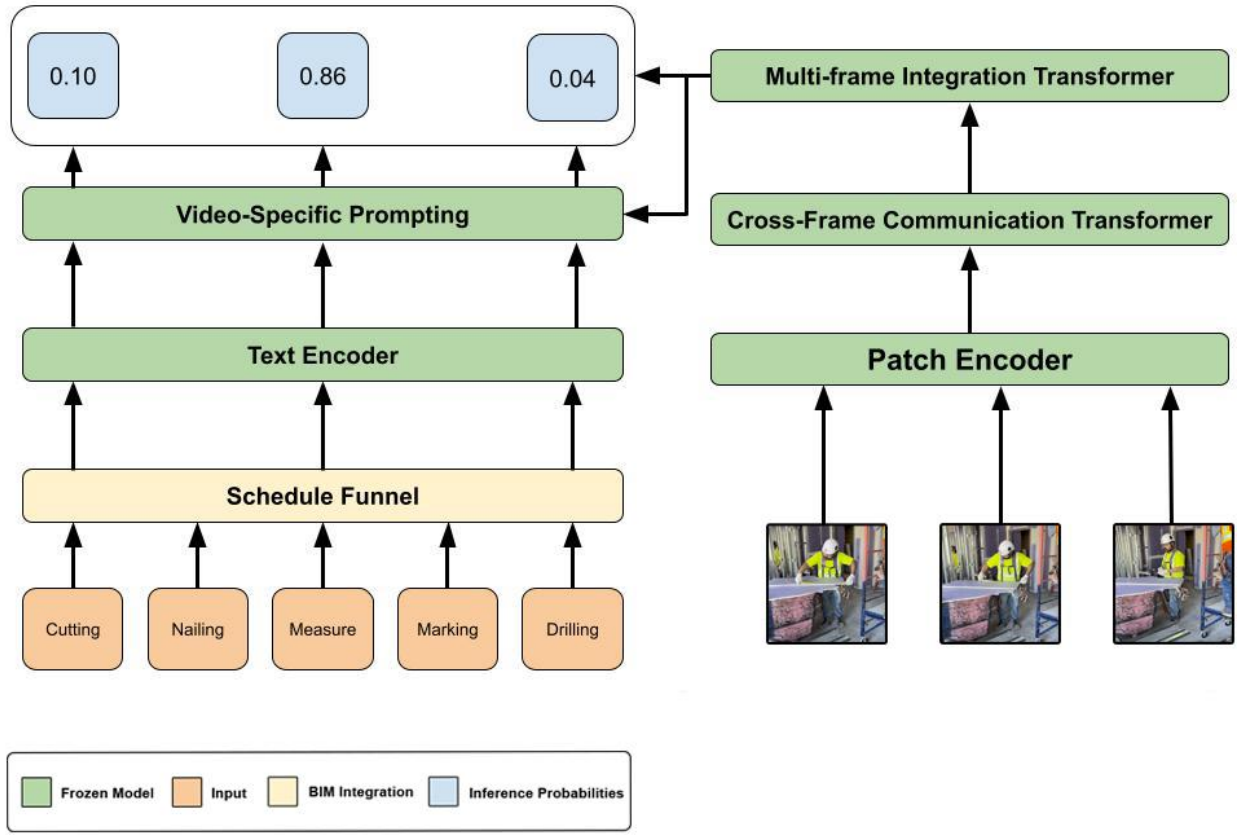

**Figure 2. X-CLIP Architecture with BIM Schedule Integration**

**Table 2. Metrics for Tasks 1 and 2 with and without BIM Integration**

| Task | BIM Integration | Accuracy | Precision | Recall | F1 Score |
|---|---|---|---|---|---|
| 1 | No | 35.71% | 0.35 | 0.36 | 0.35 |
| 1 | Yes | **57.14%** | 0.41 | 0.57 | **0.48** |
| 2 | No | 23.08% | 0.71 | 0.23 | 0.35 |
| 2 | Yes | **53.85%** | 0.59 | 0.54 | **0.57** |

As shown in Figures 3 and 4, we see an increase in maximum probabilities between both correct and all predictions for both tasks when the construction schedule is integrated. This can be

– 5 –                                                                                                      Rev.2/2024

because certain classes that share similar features are removed from the inference by the schedule's restriction. These results suggest that BIM integration can increase model confidence which can translate to more reliable and transparent field deployment.

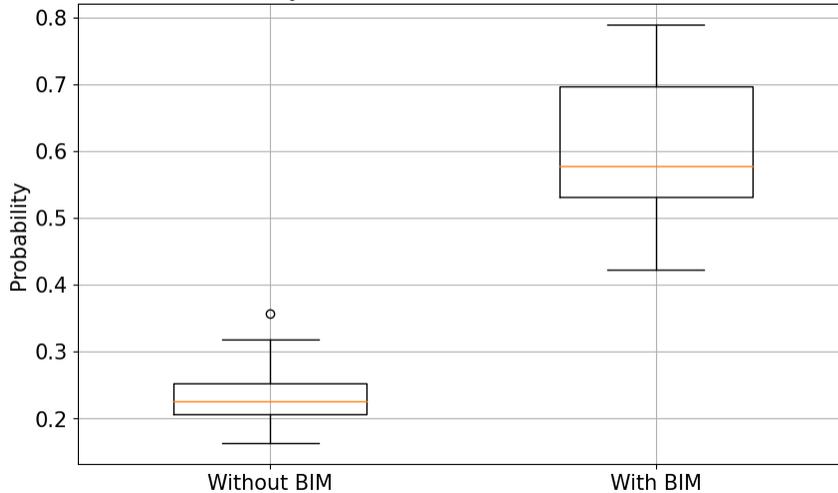

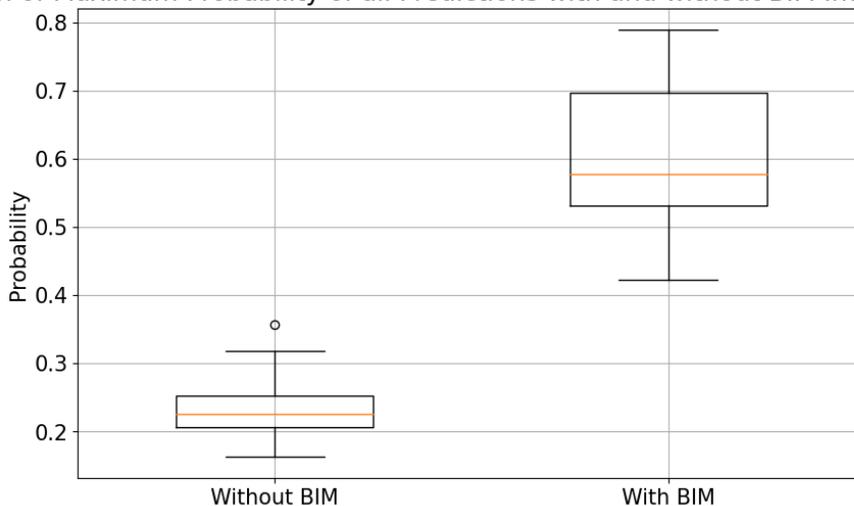

**Figure 3. Confidence Statistics for Task 1 for inferences with and without BIM schedule integration.**

**DISCUSSIONS AND LIMITATIONS**

The current schedule-to-activity transformation follows an ad libitum approach. This is a byproduct of the desired "activity level" that we decided on. Here, "activity level" pertains not to the intensity but rather to the complexity and the scale of abstraction in recognizing an activity. On the lower end of the spectrum, activities are understood in simple, fundamental terms, where even minor actions like a twitch or a joint movement are classified as activities, identifiable with minimal data. At the higher, more abstract end of the spectrum, activities are defined in more complex terms, such as performing a task like installing drywall at a construction site.

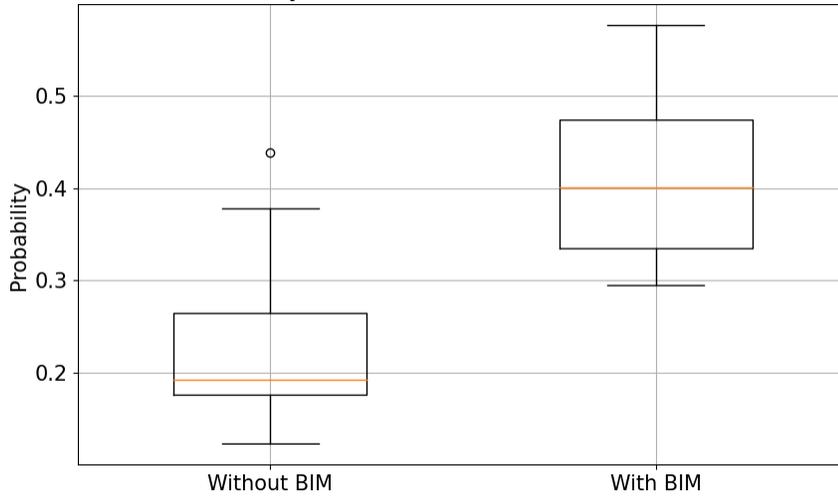

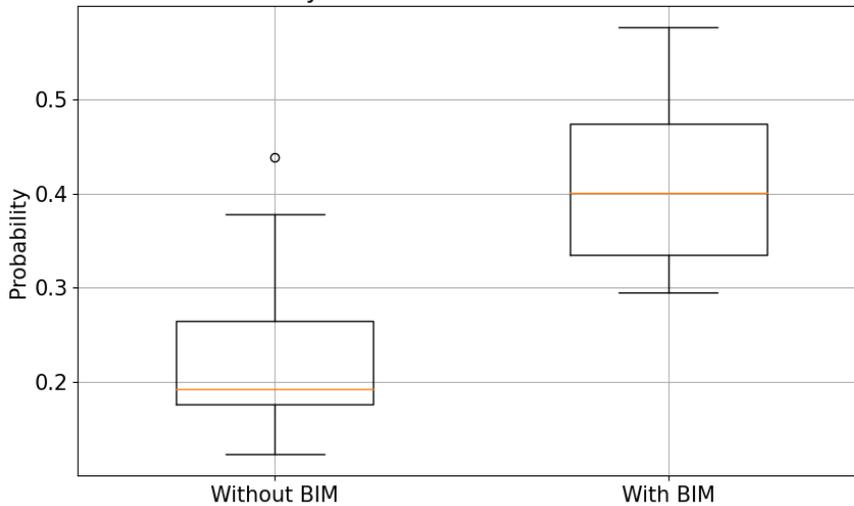

**Figure 4. Confidence Statistics for Task 2 for inferences with and without BIM schedule integration.**

The more "high-level" the activity is, the more information is needed to be able to accurately predict the activity. This increase in necessary information directly translates to an increase in video/sensory intake which will cause significant latency in processing times, which can be detrimental to real-time applications. While this choice is problem-dependent, future investigation to create a rigid framework is warranted.

One advantage that this framework provides is an increase in performance with no training and architectural changes. This is an important contribution since certain adaptations and training procedures can be too cumbersome or unfeasible. While more complex and deep model architectures can help with accuracy, the increased inference times and power requirements can prove as a detriment to practical deployments. In HRC, speed and accuracy are of paramount importance. Increasing accuracy with no increase in complexity is a significant step in the development and deployment of collaborative robots. We are currently working on improving and deploying this framework in our ongoing projects to ensure safe and effective HRC.

One limitation can be attributed to the fact that our activities might not be all-encompassing to the construction task. The selected tasks were videotaped and organized ad libitum on the job site. This can result in false predictions if an activity that has not been defined in the label space is a sub-activity of the task. Moreover, due to this framework's design, which excludes certain activities from the range of labels, inaccuracies in activity detection can arise if an activity takes place beyond the scope of possibilities predetermined by the user. This is an extreme approach that entirely removes the possibility of certain activities. A more moderate approach would be to apply penalties to specific classes given the task. However, the implications of this approach and the specifics of how penalties are applied remain unexamined, and we plan to address these topics in future works. Ideally, the model would be trained or fine-tuned with mainly construction activity data to ensure better accuracy for real-life deployments.

It is worth noting that there is currently no standard framework that translates construction tasks into activities for activity recognition inferences. However, the user has the flexibility to choose how many and which activities to include in the label space. A smaller label space could result in higher accuracy as long as the activities that occur are in that subspace, or similar activities that are seen in other tasks and not the inferred ones are penalized or removed from the label space. Nevertheless, the dynamic environment of construction can be unpredictable and affect performance if activities outside the scope of the model are observed. For future work, we plan to explore the capability and efficacy of Large Language Models as a tool for label space restriction.

**ACKNOWLEDGEMENTS**

The presented work has been supported by the U.S. National Science Foundation (NSF) CAREER Award through grant No. CMMI 2047138, as well as grant No. DUE- 1930546. The authors gratefully acknowledge the support from the NSF. Any opinions, findings, conclusions, and recommendations expressed in this paper are those of the authors and do not necessarily represent those of the NSF. We would also like to thank Clark Construction for their assistance and collaboration in this research.